\documentclass[letterpaper, 10 pt, conference]{ieeeconf}  

\IEEEoverridecommandlockouts                              

\usepackage{amsmath} 
\usepackage{caption}
\usepackage{subcaption}
\usepackage{graphicx}
\usepackage{algorithm}
\usepackage{algpseudocode}
\usepackage[utf8]{inputenc}
\usepackage{hyperref}
\usepackage{amsmath, amsfonts, amssymb}

\title{\LARGE \bf
Descriptive Model-based Learning and Control for Bipedal Locomotion
}

\author{Suraj Kumar$^{1}$, Andy Ruina$^2$
\thanks{$^{1}$ Robert Bosch Centre for Cyber Physical Systems, Indian Institute of Science; U R Rao Satellite Center,
        Bengaluru, KA, India. Email: surajk@ursc.gov.in;suraj27avionics@gmail.com%
}
\thanks{$^{2}$ Visiting Faculty, Indian Institute of Science,Bengaluru and Plaksha University, Mohali; Professor, Cornell University, USA%
}
\\
\\
}

\begin{document}

\maketitle
\thispagestyle{plain}
\pagestyle{plain}


\begin{abstract}

Bipedal balance is challenging due to its multi-phase, hybrid nature and high-dimensional state space. Traditional balance control approaches for bipedal robots rely on low-dimensional models for locomotion planning and reactive control, constraining the full robot to behave like these simplified models. This involves tracking preset reference paths for the Center of Mass and upper body obtained through low-dimensional models, often resulting in inefficient walking patterns with bent knees. However, we observe that bipedal balance is inherently low-dimensional and can be effectively described with simple state and action descriptors in a low-dimensional state space. This allows the robot's motion to evolve freely in its high-dimensional state space, only constraining its projection in the low-dimensional state space. In this work, we propose a novel control approach that avoids prescribing a low-dimensional model to the full model. Instead, our control framework uses a descriptive model with the minimum degrees of freedom necessary to maintain balance, allowing the remaining degrees of freedom to evolve freely in the high-dimensional space. This results in an efficient human-like walking gait and improved robustness.
\end{abstract}
\section{INTRODUCTION}
\label{intro}

Legged locomotion is the easiest means of mobility over a wide variety of unprepared terrains as legs can use isolated footholds to optimize support. A bipedal configuration with a human-like form factor has numerous advantages: economical walking in terms of energy efficiency, bipedal arms are well-suited to a wide variety of manipulation tasks, especially those built for humans, and mobility advantages over wheeled robots in rough terrain, cluttered indoor spaces, and environments with sparse available contact points.\\
In recent years, research focus in the field of bipedal robotics has started to shift away from simple 'lab' environments towards having humanoid robots perform useful work in the real world, as demonstrated by the DARPA Robotics Challenge (DRC) and its disaster response scenarios. Nevertheless, the existing algorithms designed for balance and locomotion control in bipedal robots have been built upon specific assumptions that are often contradicted by the actual environments where the humanoid form factor is anticipated to excel. Among all the challenges in bipedal robotics, robust balance control is perhaps the most fundamental and challenging one. \\
\begin{figure}[h]
  \centering
  \includegraphics[width=0.25\textwidth]{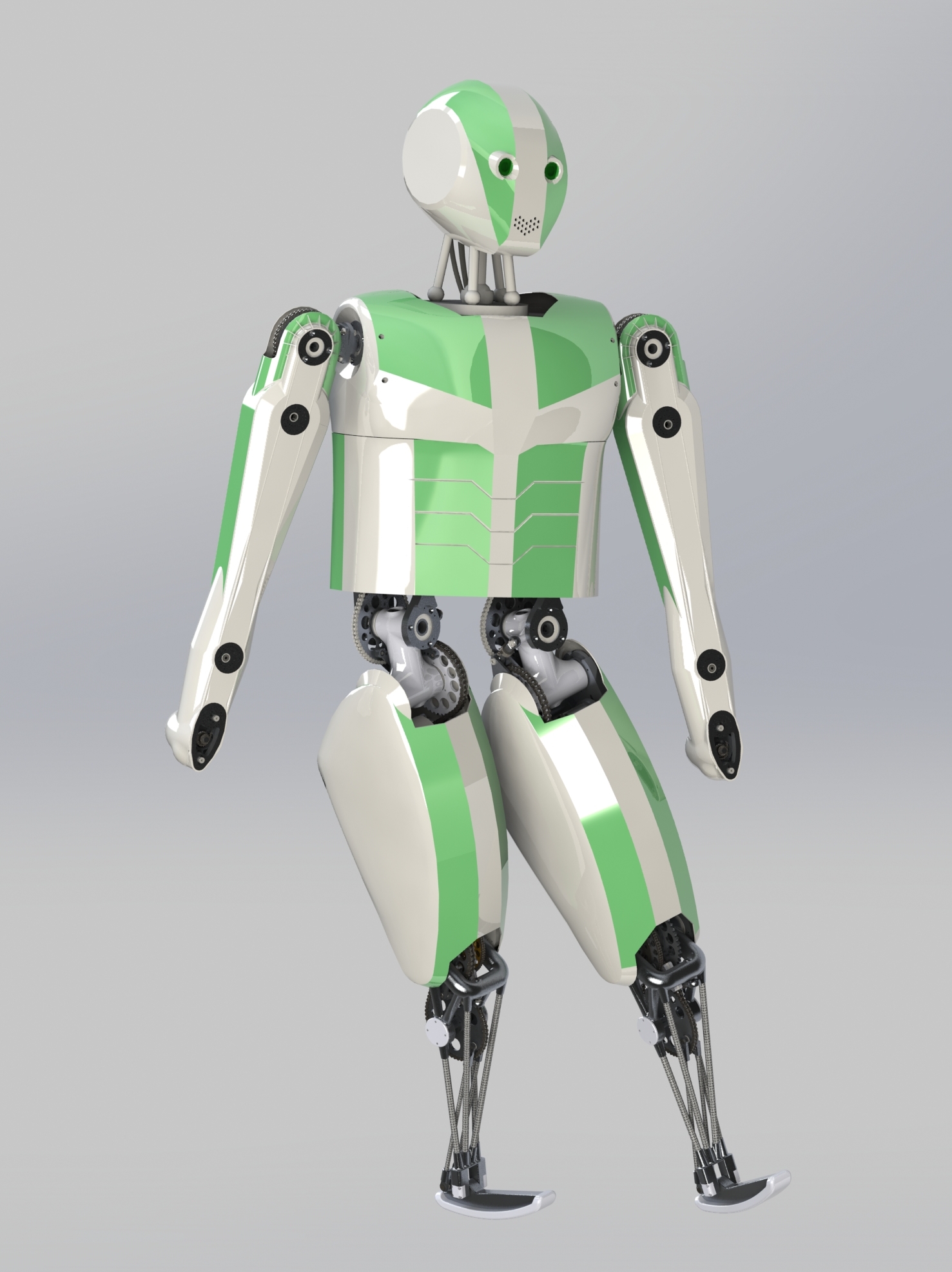}
  \caption{Ranger-Max Humanoid}
  \label{fig:ranger_max}
\end{figure}
The control problem in bipedal locomotion is studied for over two decades. Many algorithms rely on Reduced order Models (RoM) for planning and control. RoMs are low Degree-of-Freedom (DoF) representations of the original high DoF systems and capture the dominant dynamical effects of original systems. The Linear Inverted Pendulum Model (LIPM) \cite{c4,c5,c13,c14,c15} has been been widely used in walking pattern generation and foot placement control for balance owing to the analytical solution of LIPM dynamics. Task space controller based on whole body dynamics \cite{c18,c19,c6} for tracking LIPM generated reference trajectories has been successfully demonstrated in hardware \cite{c5,c6,c16}.  Inverted Pendulum Model (IP) has been used to explain energetics and boundaries of human walking in \cite{c2,c3,c8,c9} and controllers are developed based on Poincare section \cite{c10,c11,c12,c20,c21,c23,c26,c28}. Model free control methods with reduced state and action description and not utilizing dynamics has been studied and control is acquired using reinforcement learning \cite{c22,c24,c26,c27}. Recently, more expressive models such as Single Rigid Body Model (SRBM) has been adopted for balance control in Model Predictive Control(MPC) framework \cite{c17}. Nevertheless, the fundamental problem of prescription remains. This framework tracks preset reference paths using RoMs, making the full robot behave like the RoM. This requires high control bandwidth to ensure close tracking, leading to issues such as low compliance during environmental interaction, control-structure interaction with link flexible modes, and sensitivity to sensor noise. Often, the CoM dynamics from RoMs are inconsistent with the full robot dynamics under environmental uncertainties, meaning the reference trajectory does not ensure the robot's balance. Consequently, the controller has a limited basin of attraction and requires extensive gain tuning to achieve dynamically consistent CoM motion. \\
The main contribution in this work is the development of a descriptive model-based learning and control framework for bipedal locomotion that does not prescribe reference trajectories for the full robot. Bipedal walking and balance emerges from a simple projection of the full-order model into a low-dimensional space, while the remaining degrees of freedom are left free to achieve secondary tasks or improve locomotion efficiency on rough terrain. \\
The remainder of the paper is organized as follows. Section \ref{sec:prw}
provides necessary background in model reduction and control. Section \ref{sec:pbc} details components of MPC based control strategy based on LIPM prescriptive framework which doesnot assume reference foot-holds. Section \ref{sec:dbc} details components of descriptive model based control and its application in humanoid walking. 

\section{Preliminaries} \label{sec:prw}
\subsection{Dynamics of legged robots}
\label{robot_dynamics}
Legged robots are categorized as hybrid dynamical systems, featuring continuous time dynamics followed by discrete ground impact event. Mathematically, hybrid system $\Sigma$ is defined as \\ 
\begin{equation}
\label{hybrid_dyn}
\begin{aligned}
\dot{x} = f(x,u)  \quad \quad  \forall \phi(x) \neq 0, \\
x^+ = \Delta(x^-) \quad \quad   \forall \phi(x) = 0  \\
\end{aligned}
\end{equation} \\
where the continuous time dynamics is given by \(\dot{x} = f(x,u)\) with x and u representing state and control respectively;  \(\phi(x) = 0\) represents the switching or impact condition;  $\Delta$ represents the impact dynamics or impact map. \\
The bipedal robot under consideration is 18 DoF Ranger Max Humanoid as showin in figure \ref{fig:ranger_max} that weights 30 kg. Detailed specifications are listed in \cite{c34}. For a bipedal robot with $n_v$ degrees of freedom, its continuous time dynamics can be represented by standard multi-body equation with floating base given by,
\begin{equation}
\label{robot_dyn}
\begin{aligned}
    H(q)\dot{v} + C(q,v) + G(q) &= S^T_{a}\tau + J_{s}(q)^T F_{s} \\
    \dot{q} &= Q(q) v
\end{aligned}
\end{equation}
where $q \in R^{n_q}$ is the generalized joint position vector, $v \in R^{n_v}$ is the generalized joint velocity vector; $Q(q)$ is the linear map between joint position and velocity vectors; $H \in \mathbb{R}^{n_v \times n_v}$, $C \in \mathbb{R}^{n_v}$,$G \in \mathbb{R}^{n_v}$ are mass matrix, velocity product terms and gravity respectively. Here $F_s \in \mathbb{R}^{6N_{s}}$ is the stacked ground wrench corresponding to $N_{s}$ contact points and $J_s \in \mathbb{R}^{6N_{s} \times n}$ is combined contact jacobian. $\tau$ is the generalized force vector along joint axes and $S_a$ represents the actuation selection matrix. 

\subsection{Reduced order Models}
\begin{figure}[h]
  \centering
  \includegraphics[width=0.4\textwidth]{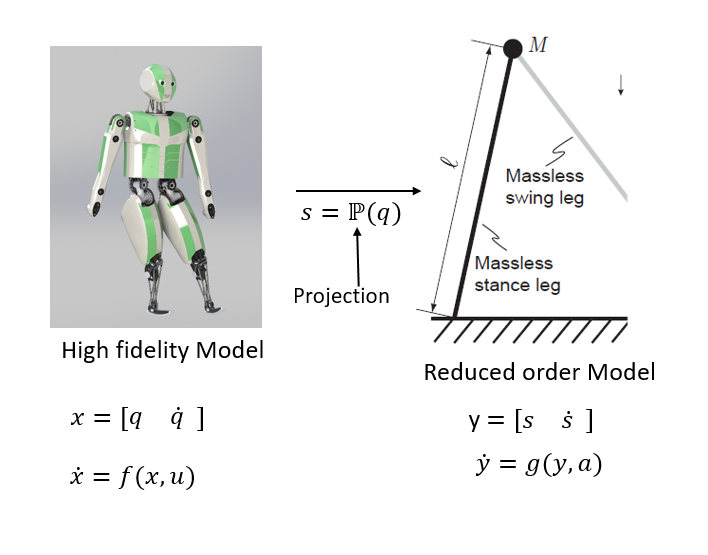}
  \caption{Full order Model vs Reduced order Model}
  \label{fig:full_vs_rom}
\end{figure}

 Let \( (s, \dot{s}, a) \) tuple represent the states and input of the Reduced Order Model (ROM). Then reduced order state, \( (s, \dot{s}) \) is defined as projection or embedding of full order states into an appropriate low dimensional vector space with \( \mathbb{P} \) representing the projection operator as shown in Figure \ref{fig:full_vs_rom}. Unlike reduced states, reduced actions are not defined as embedding of full model actions. Rather reduced actions are scalar numbers which represent aspects of walking such as foot placement, foot timing etc. essential for balance control design. Reduced order states captures the evolution aggregate motion of center of mass.

\subsection{Model Reduction Techniques in Bipedal Robots}
A RoM (also called simple model), at the core, is a standard model reduction technique used in systems and control theory. For example, linearization of a nonlinear dynamics is a model reduction technique that results in a linear model and control can be acquired for linear model and applied for non-linear system. The model reduction approach can be \textbf{prescriptive} or \textbf{descriptive} in nature. 

\subsubsection*{\textbf{Prescriptive Model reduction}}
In many model reduction approaches, control inputs are designed to force the system to follow a simple model, a 'prescriptive' approach. In bipedal robots, Linear Inverted Pendulum model (LIPM) is a prescriptive approach where-in center of mass of full robot is constraint to plane parallel to horizontal ground and upper body orientation is constrained to vertical. Major disadvantage of such prescriptive approach lies in restricting the full state trajectory in high DoF to low dimensions thus preventing the system to do anything useful while balancing itself. Any secondary motion is encoded as trajectory in either task or joint space and along with center of mass trajectory, bipedal locomotion problem is transformed into trajectory tracking problem. 

\subsubsection*{\textbf{Descriptive Model Reduction}}
 A human while walking can perform secondary tasks still maintaining balance such as leaning forward to pick up glass, kicking ball etc. Human's can do so because their overall motion is not restricted to behave like a pendulum. On the contrary, individual links and joints can move arbitrary in high dimension to achieve secondary tasks as long as their projection onto the low dimension can achieve balance. This is the spirit of descriptive map. A bipedal robot has, for balance purposes, a simple behavior that is captured with a reduced description of the system. Controller can use this simplification to achieve balance. In the descriptive approach, our interest is in generating policy that maps such reduced description of states to reduced action space without constraining the free degrees of freedom of full model. In contrast, in the ‘prescriptive’ approach to model reduction, an inner control loop forces the model to move with fewer degrees of freedom. 

\section{Prescription Based Bipedal Control} \label{sec:pbc}
\begin{figure}[h]
  \centering
  \includegraphics[width=0.5\textwidth]{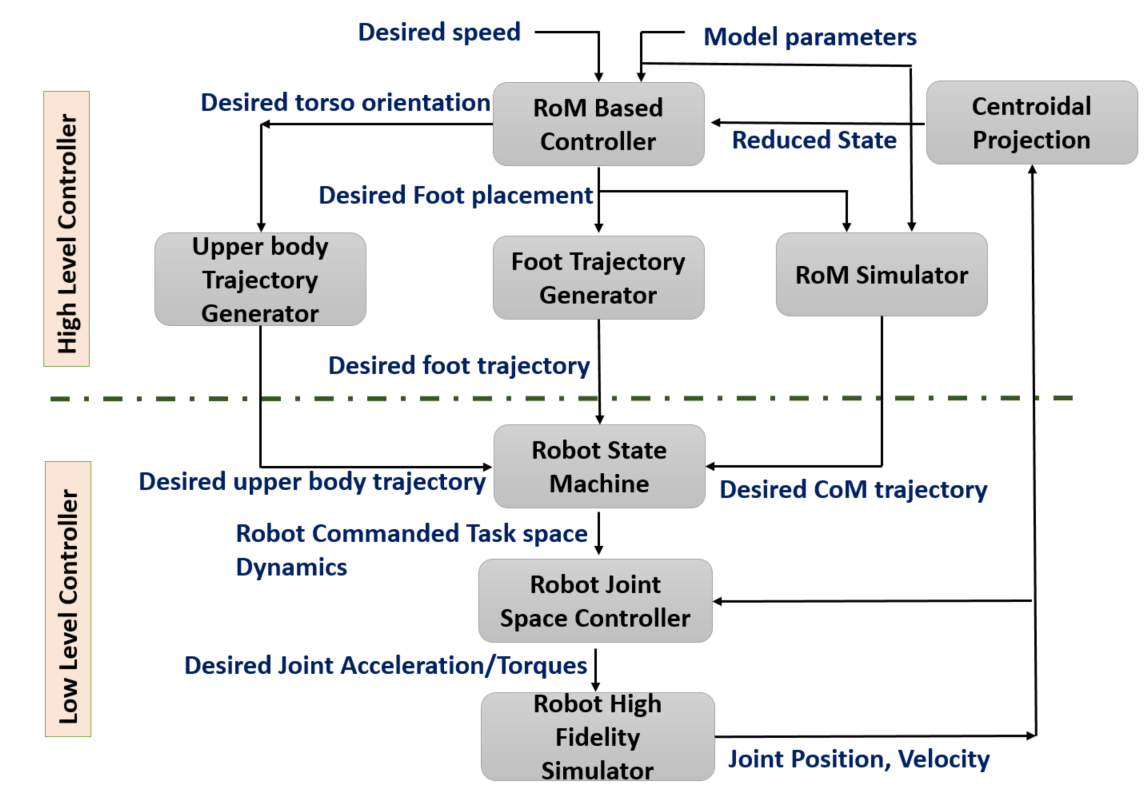}
  \caption{Prescriptive Model based Control Block Diagram}
  \label{fig:prescriptive_block_diagram}
\end{figure}

This section details components of prescriptive bipedal control framework based on Linear Inverted Pendulum Model(LIPM). The controller comprises of two layers as shown in figure \ref{fig:prescriptive_block_diagram}. High level controller uses LIPM to plan CoM trajectories and footsteps using linear discrete time Model Predictive Control. The low level controller uses optimization based task space control for tracking task space trajectories generated by high level controller.  

\subsection{LIPM Model}
In LIPM model, dynamics in sagital plane and frontal plane is decoupled and identical. 
Let $S = [x, \dot{x}]^T$ denote the CoM state and the foot placement is denoted by p. Given known timing t, future CoM state at $t_{TD}$ can then be expressed as:
\begin{equation}
\begin{aligned}
    \label{lipm_eqn}
    S(t_{TD})  = A(t_{TD})S(t) + B(t_{TD})p \\
         A(t) = 0.5\begin{bmatrix}
(e^{\omega t} + e^{-\omega t}) & \frac{(e^{\omega t} - e^{-\omega t})}{\omega} \\
\omega (e^{\omega t} - e^{-\omega t}) & (e^{\omega t} + e^{-\omega t})
\end{bmatrix}   \\
        B(t) = \begin{bmatrix}
1 - 0.5 (e^{\omega t} + e^{-\omega t}) \\
-0.5 \omega (e^{\omega t} - e^{-\omega t})
\end{bmatrix}
\end{aligned}
\end{equation}

where $w = \sqrt{g/z}$ and $t_{TD}$ denotes the touchdown timing of swing leg.

\subsection{Foot Placement}
Our first contribution is the development of a novel foot placement controller for maintaining balance while tracking desired velocity without use of pre-defined foot-holds. \\
It is assumed there is no double support phase and step time, T is considered constant to keep the formulation linear. 
Equation \ref{lipm_eqn} can be propagated for N steps resulting in N linear equations given step time T as fixed parameter:
\begin{equation}
\begin{aligned}
    \label{lipm_n_eqn}
    S_1 & =  A(t_{TD})S_0 + B(t_{TD})p_0 \\
    S_2 & =  A(T)S_1 +  B(T)p_1 \\
    S_3 & =  A(T)S_2 +  B(T)p_2 \\
    S_4 & =  A(T)S_3 +  B(T)p_3 \\
    & \vdots \\
    S_N & =  A(T)S_{N-1} + B(T)p_{N-1}
\end{aligned}
\end{equation}
Current support foot position, $p_{0}$ is fixed and cannot be changed but future foot placements can be optimized to achieved balance.
The foot placement controller can be written as Quadratic program:
\begin{equation}
\begin{aligned}    
    \underset{p_1,p_2,...,p_N}{\text{minimize}} 
    & \sum_{k=1}^{N-1} ((S_{k+1} - S^*)^T Q (S_{k+1}-S^*) \\
    & + (p_{k+1}-p_k)^T R (p_{k+1}-p_k))  
\end{aligned}
\end{equation}
subject to dynamics constraints \ref{lipm_n_eqn}

\subsection{Task Space Control}
\begin{figure}[h]
  \centering
  \includegraphics[width=0.5\textwidth]{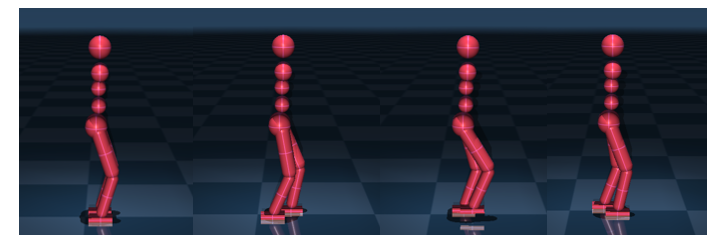}
  \caption{LIPM Walking Gait}
  \label{fig:lipm_walking}
\end{figure}

Torso pose, foot step location, CoM dynamics trajectory generated by MPC as shown in figure \ref{fig:prescriptive_block_diagram} can be tracked using Task Space Controller(TSC). TSC generates desired joint torques needed to track reference trajectories under whole body dynamics constraints along with contact and unilateral constraints as follows
\begin{equation}
\begin{aligned}
    \label{tsc_problem}
    & \underset{\ddot{q},\tau,F_{s}}{\text{minimize}} 
    & & ||A_{t}(q)\ddot{q} + \dot{A_{t}}(q) \dot{q} - \dot{r}_{t,c}||^2_{Q_{t}} + ||\tau||^2_{R_{\tau}} + ||F_{s}||^2_{R_{\lambda}} \\
    & \text{subject to} 
    & & H(q)\ddot{q} + C(q,\dot{q}) + G(q) = S^T_{a}\tau + J_{s}(q)^T F_{s} \\ 
    & & & J_{s}(q) \ddot{q} + \dot{J}_{s}(q)\dot{q} = 0 \\
    & & & F_{s} \in P \\ 
    & & & |\tau| < \tau_{max}
\end{aligned}
\end{equation}
$A_{t}(q)$ is referred to as generalized jacobian relating joint velocities to task and can be computed from robot kinematics; $P$ denotes the contact constraint set comprising of unilateral force constraints, friction pyramid constraints as well as ZMP constraints; $\dot{r}_{t,c}$ is the commanded task space dynamics imposed as linear PD control for tracking desired reference trajectory for the associated task as:
\begin{equation}
    \label{task_dyn_lin_form}
    \dot{r}_{t,c} = \ddot{p}^d + K_{P}(p^d - p) + K_{D}(\dot{p}^d - \dot{p})
\end{equation}
Figure \ref{fig:prescriptive_block_diagram} shows standard implementation block diagram of LIPM locomotion controller. Foot placement controller computes next foot placement based on desired walking speed which is fed to foot trajectory generator module that computes desired foot trajectory from current foot position to target foot position. Foot placement and model parameters are used by RoM simulator to compute center of mass trajectory. In LIPM, upper body orientation is regulated to vertical as prescribed by the LIPM equation of motion. More expressive models such as single rigid body model allows modulation of torso orientation for maintaining balance. This is fed to upper body trajectory generator module to generate torso orientation trajectory. Robot state machine then selects appropriate task dynamics for the foot, CoM and upper body to continuously track the RoM-CoM dynamics and realize the desired foot touchdown locations. Joint space controller then selects whole-body joint torques/joint accelerations at real-time to realize these desired tasks using task space controller. Figure \ref{fig:lipm_walking} shows stable walking gait obtained from this implementation.
\section{Description Based Bipedal Control} \label{sec:dbc}
Humans walk autonomously without planning joint's motion before each step. Instead, our central nervous system regulates key parameters like ground clearance, forward velocity, swing foot placement relative to the center of mass, and postural configuration to minimize walking effort. In bipedal systems, controlling velocity is challenging because the degrees of freedom contributing to velocity vectors are under-actuated in continuous dynamics point of view. Prescriptive control method as described in section \ref{sec:pbc} embeds velocity control in a prescribed center of mass trajectory, tracked using joint torques, often resulting in inefficient walking patterns with bent knees. Essentially, we need a control framework that meets essential requirements without prescribing specific motions. This section details the components of descriptive model based control framework that achieves natural and efficient bipedal locomotion by focusing on such key parameters without prescribing desired motions. 

\subsection{Descriptive Inverted Pendulum Model}
\begin{figure}[h]
  \centering
  \includegraphics[width=0.45\textwidth]{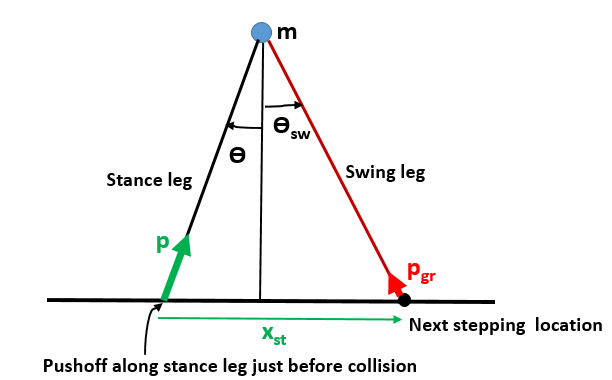}
  \caption{IP Model}
  \label{fig:ip}
\end{figure}

\begin{figure}[h]
  \centering
  \includegraphics[width=0.5\textwidth]{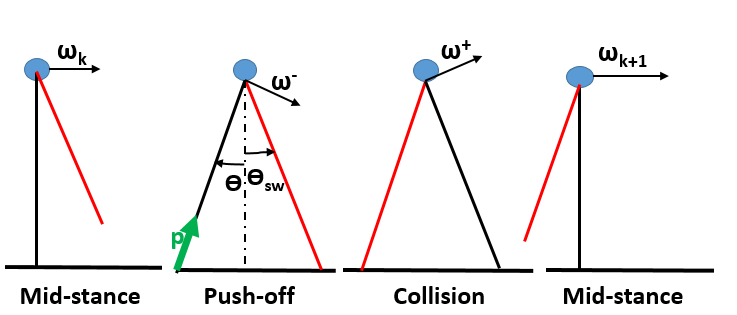}
  \caption{One step of IP model}
  \label{fig:ip_step}
\end{figure}

We first describe Inverted Pendulum (IP) model based on powered dynamic walker inspired by human walking gait cycle and is descriptive in nature.  It comprises of a point-mass hip with mass-less legs. The model is depicted in Figure \ref{fig:ip}. The system has one dynamic variable (the stance leg angle $\theta$) and two states (namely $\theta$ and $\dot{\theta}$). There are two control parameters per each step: the step length $x_{st}$, which determines the next stepping location and time, and the push-off impulse p, applied along the stance leg just before the foot strike. Owing to the fact that, ankle torques and body distortion based balance strategies have substantial less effect in balance as compared to recovery steps and push-off, IP model keeps degrees of freedom associated with upper body distortion and ankle torques (except for at push-off) free and use only stepping and push-off to control the balance. The impulsive push-off is meant to be a proxy for the energy injected into the walking motion by the extension of the trailing leg (ankle extension). 
A walking step comprises of four distinct phases, starting from mid-stance, as illustrated in figure \ref{fig:ip_step} : swing-down to the step angle ($\theta_{sw}$) followed by an impulsive push-off (p); then heel-strike and leg-switch; and finally a swing-up to mid-stance.
 \\
The angular rotation rate immediately before push-off is obtained by conservation of energy as follows:
\begin{equation}
\label{ip_eq1}
(\omega^{-})^{2} = (\omega_{k})^{2} + \frac{2g}{l}(1 - \cos{\theta_{sw}})
\end{equation} \\
Next, a push-off impulse is applied to the point-mass along the stance leg, changing the hip’s velocity vector. After that, the swing leg becomes the new stance leg as it collides with
the ground, exerting another impulse on the point-mass hip. The composition of these two collisions, governed by angular momentum balance about the new stance foot, yields the
rotational speed of the new stance leg:
\begin{equation}
\label{ip_impact_eq}
\omega^{+} = \omega^{-}\cos{2\theta_{sw}} - \frac{p}{ml}\sin(2\theta_{sw}) 
\end{equation}
After the two collisions, the stance leg swings up to the next mid-stance, again ruled by conservation of energy,
\begin{equation}
\label{ip_eqn2}
(\omega_{k+1})^{2} = (\omega^{+})^{2} - \frac{2g}{l}(1 - \cos{\theta_{sw}})
\end{equation} 

\subsection{Foot-placement Control}
IP model is used to describe a Markov Decision Process (MDP) and the resultant MDP is solved via Reinforcement Learning (RL). Lets denote the state and action at $k^{th}$ step by $s_{k}$ and $a_{k}$. The cost associated associated with a control policy $\pi$ is defined in terms of reward as
\begin{equation}
\label{cost_fn1}
J_{\pi}(s) = \mathbb{E}_{\pi}(\sum_{k=0}^{\infty}\gamma^{k}r(s_{k},a_{k},s_{k+1})|\pi)
\end{equation}

\begin{equation}
\label{cost_fn1}
\begin{aligned}
r(s_{k},a_{k},s_{k+1}) = c_{1} \exp(-(v-v_{des})^{\alpha_1}) +  \\ 
                        c_{2} \exp(-(\frac{d}{T})^{\alpha_{2}}) +  \\
                        c_{3} \exp(-(pv)^{\alpha_{3}})  
\end{aligned}
\end{equation}
The controller is to minimize an infinite horizon cost function composed of three terms; cost associated with deviation from desired orbital velocity, swing cost associated with hip actuation constraints and energetic costs associated with work done by robot in push-off.

\subsection{Velocity Control}
\begin{figure}[h]
  \centering
  \includegraphics[width=0.45\textwidth]{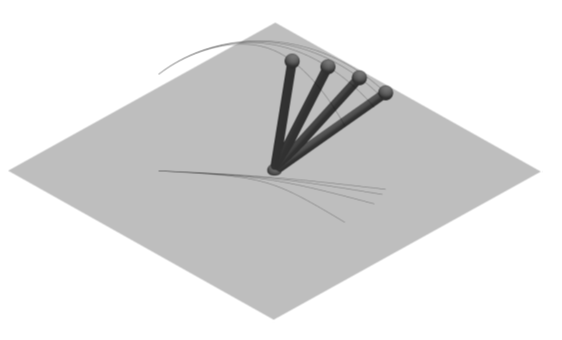}
  \caption{Walking Orbits}
  \label{fig:orbits}
\end{figure}

The balance control model uses impulse as a proxy for the energy needed to maintain desired speed. However, impulse-based inputs cannot directly actuate ankle joints in the full model. Therefore, we control the center of mass velocity explicitely using ankle torques via orbital parameter regulation. In celestial mechanics, satellites move along static orbit described by static set of parameters during periods of unforced motion. An orbital description of state consists of two parts: the orbit an object is traveling on, and the location of the object along this orbit. In context of walking, the unforced motion of the robot during the swing phase can be thought of as an orbit as shown in figure \ref{fig:orbits}. In contrast to the orbits of celestial mechanics, walking orbits are not periodic and switches from orbit to orbit during stance change to avoid falling. These orbits are defined by static parameters, $o(t)$ namely energy, momentum and location of orbit in space. So velocity regulation controller is transformed in orbit transfer problem and controller learns a policy from orbital state to ankle torque given as $ \tau_{st-ankle} = \Pi_{v}(o(t))$. A linear policy is learnt for velocity regulation in our simulation and analysis. The controller is to minimize the error in orbital states, $o(t)$ from the desired orbital states and ankle torque.

\begin{figure}[h]
  \centering
  \includegraphics[width=0.45\textwidth]{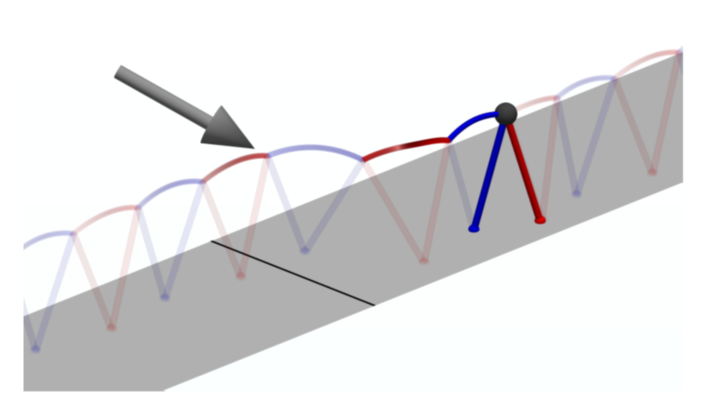}
  \caption{IP Walking under disturbance}
  \label{fig:ip_walking}
\end{figure}

\begin{figure}[h]
  \centering
  \includegraphics[width=0.5\textwidth]{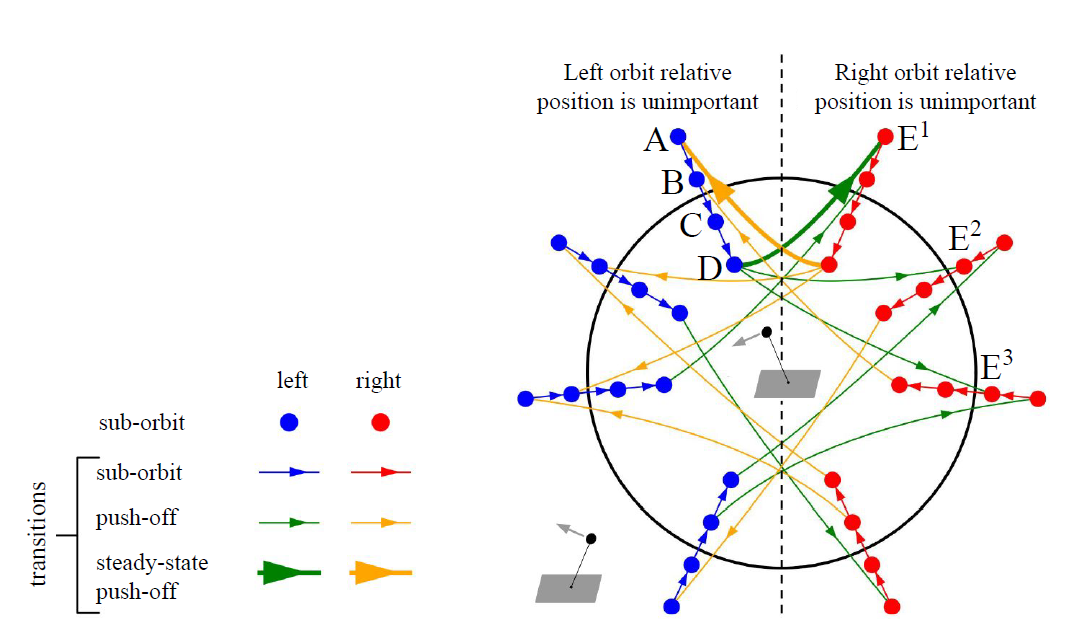}
  \caption{Balance Control Visualization}
  \label{fig:ip_control_visualization}
\end{figure}

Figure \ref{fig:ip_walking} shows IP walking under disturbance. IP recovers when subjected to a disturbance when it crosses the black line. Left orbits are
shown in blue, and right orbits are shown in red. Balance is achieved by appropriate switching of the orbits from disturbed state to desired state. Figure \ref{fig:ip_control_visualization} shows visualization of IP balance control problem. To understand
this figure, begin at the node marked A. Node A is a left sub-orbit and represents a set of orbital parameters and phase bounds. This node is outside the large black circle, so the pendulum is traveling upwards. Because the pendulum is traveling upwards, stance change is not possible. Node A connects to node B. This connection is marked with a blue arrow. This connection represents a transition from one sub-orbit due to the next due to the passage of time. Node B connects to node C. This node is inside the large black circle, so the pendulum is traveling downwards. Because the pendulum is traveling downwards, stance change is possible
in principle, but for this particular state, no exact transitions to other sub orbits exist. Therefore node C must transition to node D. Stance change is possible at Node D, to either node E1, node E2, or node E3. These left to right stance change transitions, and are marked with green arrows. The steady-state stance change transition is the one to node E1, and is marked with a thick green arrow. 

\subsection{Humanoid 2D Control}
\begin{figure}[h]
  \centering
  \includegraphics[width=0.5\textwidth]{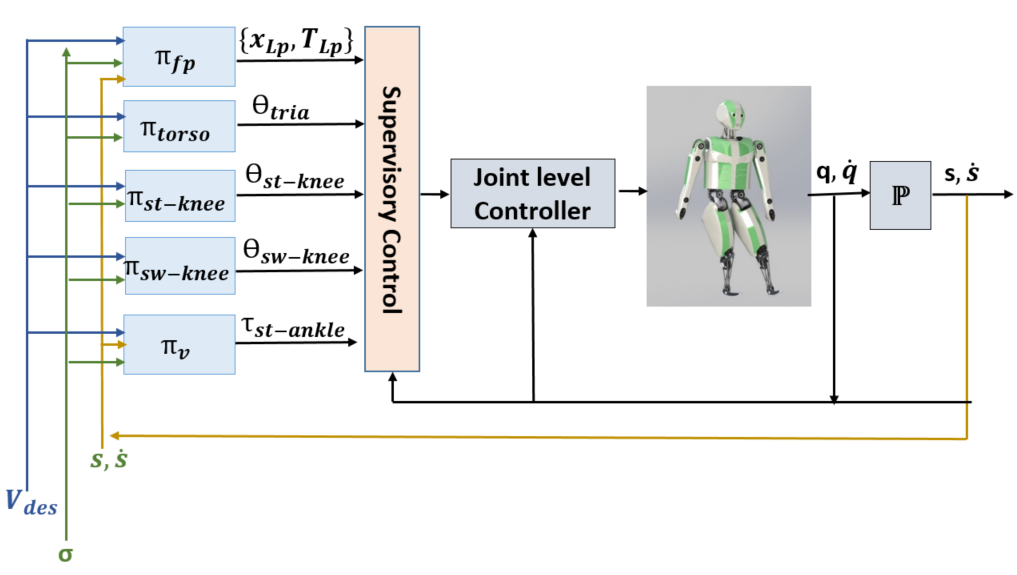}
  \caption{Humanoid Control Block Diagram}
  \label{fig:hum_des_block_diagram}
\end{figure}

We now describe the control architecture used for realizing stable and robust walking based on descriptive IP model. The controller architecture consists of 4 main parts as shown in figure \ref{fig:hum_des_block_diagram}: \textbf{Balance control} that defines balance strategies; \textbf{Posture control} that defines control strategies for free degrees of freedom associated with specific choice of descriptive model; \textbf{Supervisory Control} that manages and coordinate among different controllers; \textbf{Joint level control} that computes desired joint acceleration/torques to realize actions defined by balance and posture controller. We now explain the details of each controller and its design procedure.
\subsubsection*{\textbf{Balance Control}}
Balance control is derived from descriptive IP model and it comprises of foot placement control and velocity regulation control as described above. 
\subsubsection*{\textbf{Posture control}}
The bipedal system under consideration is an 18-DOF system in 3D, which reduces to a 9-DOF system when constrained to the sagittal plane. In this 9-DOF 2D system, the robot's state is described by the joint states in 12 dimensions, excluding the 6 dimensions associated with the floating base state, meaning there are 12 functions of time. The balance controller provides only two outputs: the foot location along with the time of impact and ankle torques during push-off . This implies that the balance controller only constrains the swing foot's position at one specific moment in the walking cycle. Consequently, the 12 functions of time corresponding to the joint states remain unconstrained by the balance controller. Therefore, we define posture control, which manages these degrees of freedom to achieve stable walking and enhance robustness against environmental disturbances. The states controlled by posture control are independent of the actual location of robot. \\ 
Given the numerous degrees of freedom in posture states, we have selected the following set of variables for control.
\begin{enumerate}
    \item Desired torso orientation along an inertial frame 
    \item Robot stance leg length
    \item Robot swing knee touchdown angle
\end{enumerate}
These parameters are selected to control only key parameters of walking without prescribing a desired motion.When properly designed, it results in robust dynamic walking. We now explain the design criteria for generating control policy for each of these variables.

\subsubsection*{\textbf{Torso Policy}}
For planar flat terrain walking, the torso can be regulated to a vertical orientation. Generally, it is desirable for the robot to have an inertial fixed reference axis, denoted as $\Theta_{\text{tria}}$, along which the upper body aligns during dynamic walking. Varying $\Theta_{\text{tria}}$ optimizes forward lean for higher speeds on level ground, provides extra gravity moment by keeping the CoM ahead of the CoP, aids braking on down-slopes, and improves standing balance stability by maintaining the CoM projection within the support polygon. Intelligent control of $\Theta_{\text{tria}}$, enhances postural stability and robustness against disturbances.  The policy takes the form: 
\begin{equation}
    \Theta_{tria} = \Pi_{Torso}(v_{des},\sigma)
\end{equation}
where $\sigma$ denote the local slope. 

\subsubsection*{\textbf{Stance Leg Length Policy}}
Stance leg length actively controls the height of CoM from the ground and provides braking and forward thrusting during bipedal locomotion in an implicit manner. For level walking, this variable is controlled to near straight leg posture that offers energetic benefits, requiring far less knee torque. For down-slopes, leg length control facilitates hip drooping so that swing foot can land on the ground surface which lies below the current stance foot level. Let $\Theta_{st-knee}^0$ denote the stance knee joint desired orientation. The policy associated with $\Theta_{st-knee}^0$ takes the form:
\begin{equation}
    \Theta_{st-knee}^0 = \Pi_{st-knee}(v_{des},\sigma)
\end{equation}
where $\sigma$ denote the local slope.
\subsubsection*{\textbf{Swing Knee Touchdown Angle Policy}}
The swing knee joint angle at the time of heel touchdown is an important parameter which affects the gait stability. An excess bent knee during heel strike is not suitable for regulating the height of CoM in the subsequent stance phase whereas a near straight knee during heel strike leads to excess velocity reduction as well as kinematic singular state. For level walking and downslopes, near straight leg orientation during heel strike as observed in humans is preferred where as for up-slopes, intelligent manipulation of knee angle provides gait stability. Let $\Theta_{sw-knee}^0$ denote the stance knee joint desired orientation. The policy associated with $\Theta_{sw-knee}^0$ takes the form:
\begin{equation}
    \Theta_{sw-knee}^0 = \Pi_{sw-knee}(v_{des},\sigma)
\end{equation}
where $\sigma$ denote the local slope. 

The policies $\Pi_{Torso}, \Pi_{st-knee}, \Pi_{sw-knee}$ are obtained through simulation based optimization directly from data obtained from full model.

\subsubsection*{\textbf{Supervisory Control}}
Humanoid locomotion is an event driven cyclic process as robot moves through ground feet interaction during walking. The walking gait is divided into distinct event phases based on the contact between the feet and the ground. These phases are: single support phase, double support phase, and flight phase, which correspond to contact with one foot, both feet, or no feet on the ground, respectively. Supervisory control implements state machine based on ground feet interaction and coordinates among different controllers. 
\subsubsection*{\textbf{Joint level controller}} : Actions from high level controllers such as balance and posture control is realized using a PD controller on joint position and velocity. 
\begin{figure}[h]
  \centering
  \includegraphics[width=0.45\textwidth]{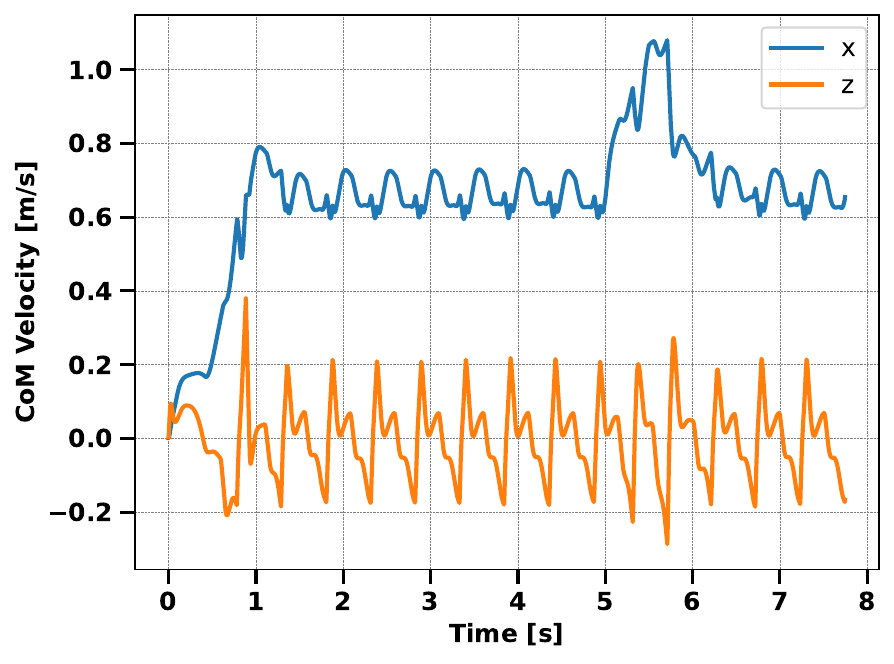}
  \caption{Humanoid's CoM velocity}
  \label{fig:com_vel}
\end{figure}
\begin{figure}[h]
  \centering
  \includegraphics[width=0.5\textwidth]{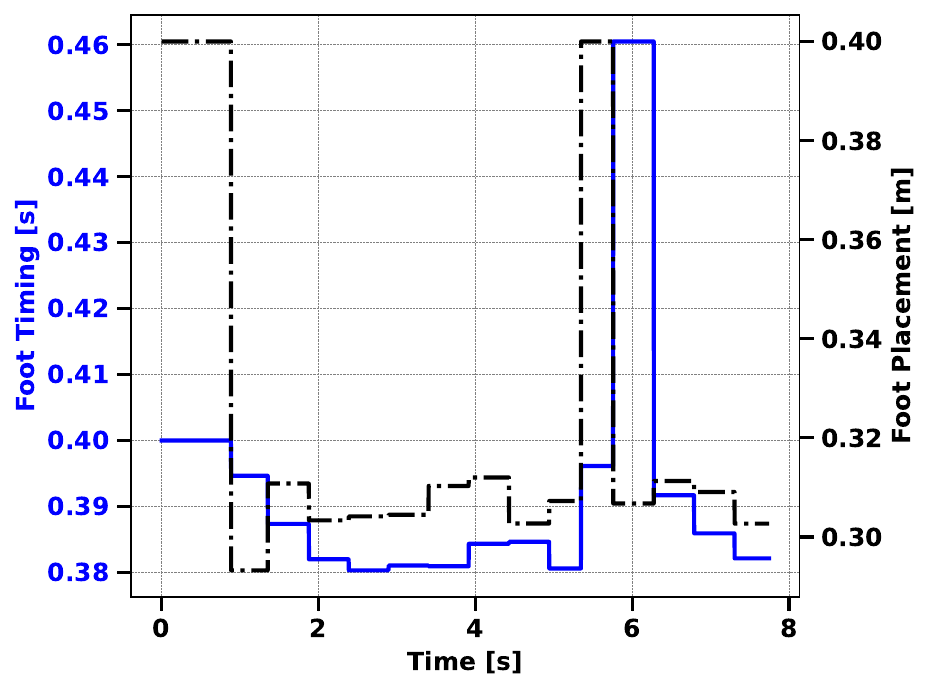}
  \caption{Foot placement}
  \label{fig:foot_placement}
\end{figure}
\subsection{Results}
The stability and agility of the proposed control scheme are demonstrated based on simulation
experiments conducted using Ranger Max humanoid robot constraint to sagital plane defined by forward motion in XZ plane. MuJoCo\cite{c35} is used for multi-body dynamics simulation. Simulation experiments are conducted for different walking speed ranging from 0.2 m/s to 2 m/s and for ground slopes ranging from -10 deg to 10 deg and proposed controller demonstrate stable walking in each case. We present simulation result of one sample scenario under external disturbance to demonstrate the stability of proposed control framework. Video demonstration of stable walking for each individual cases can be accessed \href{https://indianinstituteofscience-my.sharepoint.com/:f:/g/personal/surajkumar13_iisc_ac_in/EkXLkJgIW5xEurke5K4fNQwBKSWomeCJjpy5LHOllQT_qg?e=OM3vR9} {here}.
\begin{figure*}[h]
  \centering
  \includegraphics[width=0.9\textwidth]{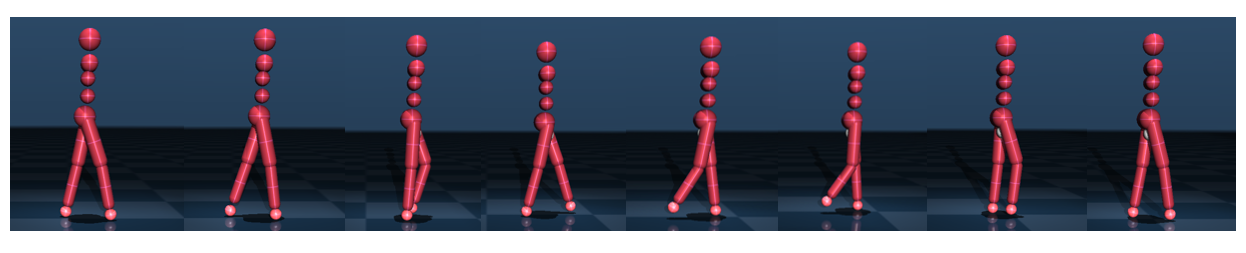}
  \caption{Descriptive IP Walking Gait Cycle for flat terrain}
  \label{fig:ipm_walking}
\end{figure*}
\begin{figure*}[h]
  \centering
  \includegraphics[width=0.9\textwidth]{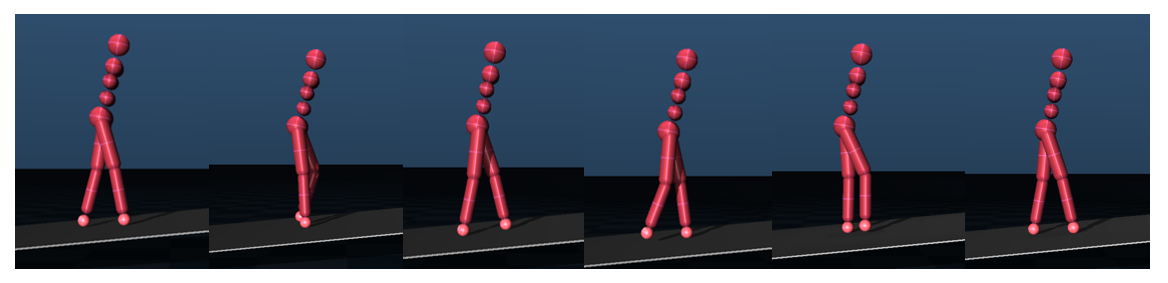}
  \caption{Descriptive IP Walking Gait Cycle for 10 deg up-slope}
  \label{fig:ipm_walking_upslope}
\end{figure*}
\begin{figure*}[h]
  \centering
  \includegraphics[width=0.9\textwidth]{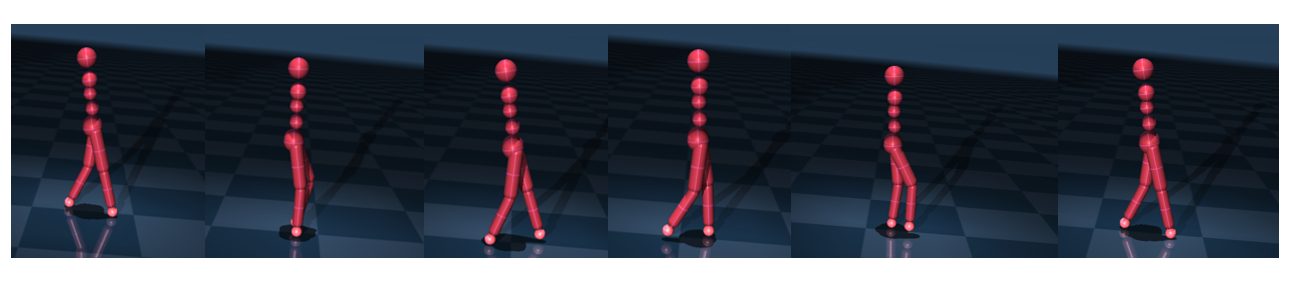}
  \caption{Descriptive IP Walking Gait Cycle for 10 deg down-slope}
  \label{fig:ipm_walking_downslope}
\end{figure*}
In this simulation, robot starts from rest and commanded desired velocity of 0.65 m/s. Figure \ref{fig:com_vel} shows the velocity of robot. It takes 3 steps to reach the desired commanded velocity. At 5s in simulation, push disturbance of 100N for 0.2s is applied on robot which disturbs robot velocity and increases to 1.2 m/s. Controller takes reactive steps shown in figure \ref{fig:foot_placement} and absorbs the disturbance in 3 steps and reaches steady state desired velocity. 

Figure \ref{fig:ipm_walking}, \ref{fig:ipm_walking_upslope}, \ref{fig:ipm_walking_downslope} shows snapshot of stable walking cycle obtained from the controller on flat, up-slope and down-slope terrain respectively.

\section{CONCLUSION AND FUTURE WORKS} \label{fw}
We have presented a novel control framework for bipedal locomotion based on descriptive model reduction technique that is robust and doesn't rely on tracking preset reference paths based on simple model. Resultant controller uses minimum degrees of freedom for balance and keeps remaining degrees of freedom in humanoid free. We identified key parameters for control among the free degrees of freedom and designed controller for these parameters directly from simulation based optimization. Resultant controller can regulate a desired walking speed while preventing falls due to external disturbances upto velocity perturbation of 2m/s. We validated the control framework on 2D humanoid model and demonstrated stable walking for different range of scenarios including up and down slopes upto 10 deg. \\
In the future, we aim to extend walking controller to rough terrains, stairs and extend this control framework for 3D humanoid and demonstrate in hardware.

\section{ACKNOWLEDGEMENTS}
This work was supported by Robert Bosch Centre for Cyber Physical Systems, IISc





\end{document}